\pdfoutput=1
\documentclass[11pt]{article}

\usepackage[final]{EMNLP2023}

\usepackage{times}
\usepackage{latexsym}
\usepackage[T1]{fontenc}
\usepackage[utf8]{inputenc}
\usepackage{microtype}
\usepackage{inconsolata}

\usepackage{hyperref}
\usepackage{graphicx}
\usepackage{multirow}
\usepackage{colortbl}
\usepackage{tikz}
\usepackage{adjustbox}
\usepackage[normalem]{ulem}
\usepackage{mathbbol}
\usepackage{amsmath}
\usepackage{bm}
\usepackage{tcolorbox}
\usepackage{textcomp}
\usepackage{makecell} 
\usepackage{pgfplots}
\usetikzlibrary{patterns}
\pgfplotsset{width=10cm,compat=1.9}

\definecolor{rred}{RGB}{255, 177, 176}
\definecolor{oorange}{RGB}{255, 223, 190}
\definecolor{yyellow}{RGB}{255, 255, 191}
\definecolor{ggreen}{RGB}{180, 240, 167}
\definecolor{bblue}{RGB}{169, 209, 247}

\title{Can Large Language Models Capture Dissenting Human Voices?}

\author{
      Noah Lee\thanks{\space{ } Equal contribution} \\
      KAIST AI \\
      \texttt{noah.lee@kaist.ac.kr} \\\And
      Na Min An\footnotemark[1] \\
      KAIST AI \\
      \texttt{naminan@kaist.ac.kr} \\\And
      James Thorne \\ %$^*$ \\
      KAIST AI \\
      \texttt{thorne@kaist.ac.kr}}

\begin{document}
\maketitle

\begin{abstract}
Large language models (LLMs) have shown impressive achievements in solving a broad range of tasks. Augmented by instruction fine-tuning, LLMs have also been shown to generalize in zero-shot settings as well. However, whether LLMs closely align with the human disagreement distribution has not been well-studied, especially within the scope of natural language inference (NLI). In this paper, we evaluate the performance and alignment of LLM distribution with humans using two different techniques to estimate the multinomial distribution: Monte Carlo Estimation (MCE) and Log Probability Estimation (LPE). As a result, we show LLMs exhibit limited ability in solving NLI tasks and simultaneously fail to capture human disagreement distribution. The inference and human alignment performances plunge even further on data samples with high human disagreement levels, raising concerns about their natural language understanding (NLU) ability and their representativeness to a larger human population.\footnote{The source code for the experiments is available at \href{https://github.com/xfactlab/emnlp2023-LLM-Disagreement}{https://github.com/xfactlab/emnlp2023-LLM-Disagreement}.}
\end{abstract}

\section{Introduction}
Natural language inference (NLI) has long served as a fundamental testbed to evaluate the ability of a model to recognize entailment and capture plausible inference relations between pairs of sentences \citep{dagan2006pascal,bowman2015large, williams-etal-2018-broad}. When constructing datasets, conventional processes result in a single label per instance even if multiple annotators contribute, which limits the full representation of diverse opinions that might arise in a larger human population. Thus, recent datasets have become more attentive to incorporating multiple interpretations \cite{pavlick2019inherent, nie2020can, glockner2023ambifc} to capture dissenting human opinions.

\hspace{1mm}

\begin{figure}[t!]
\centerline{\includegraphics[width=6.8cm]{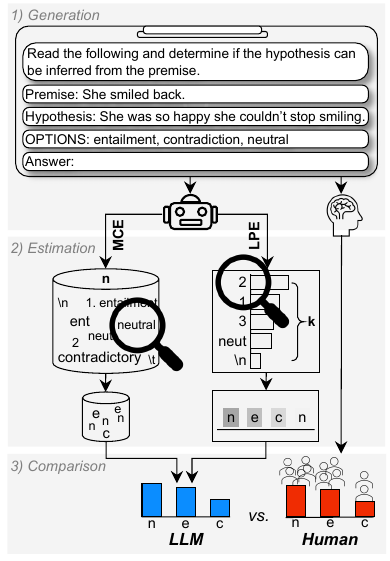}}
\vspace*{0mm}
\caption{The Proposed LLM Distribution Estimation Techniques, MCE and LPE. We estimate LLM disagreement with either MCE or LPE utilizing generated LLM outputs and compare the estimated LLM distribution with human disagreement distribution.}
\label{fig:fig1}
\end{figure} 

Meanwhile, instruction fine-tuning large language models (LLMs) has elicited remarkable generalizability to diverse unseen tasks \cite{zhao2023survey}. Not only can they generate free-form texts, but they can also select one answer from multiple options given in the input prompt. However, while many works study user interaction and conversational usage \cite{liang2022holistic}, limited works evaluate these instruction-following LLMs on a foundational NLI task. Therefore, we aim to answer the following questions: Can LLMs capture dissenting voices that naturally arise in the dataset? Are LLMs representative of the voices of the annotators in inference tasks?

With this in mind, we jointly assess on a number of instruction-following LLMs, Flan-T5 \cite{chung2022scaling}, Flan-UL2 \cite{tay2022ul2}, OPT-IML-Max \cite{iyer2022opt}, and GPT-3 \cite{ouyang2022training}, on their performance on human opinion distribution datasets - ChaosNLI \cite{nie2020can} and PK2019 \citep{pavlick2019inherent}. For the process of using the model output distribution as an estimate of human disagreement distribution, we offer novel estimation methods: Monte Carlo Estimation (MCE) and Log Probability Estimation (LPE) (Figure \ref{fig:fig1}).

We find that the state-of-the-art GPT-3 model does not outperform smaller models such as fine-tuned BERT \cite{devlin2018bert} and partially fine-tuned Flan-T5-XXL in solving inference problems. Furthermore, it yields higher Jensen-Shannon Distance (JSD) \cite{endres2003new} and Distribution Calibration Error (DCE) \cite{baan2022stop} than BERT for the ChaosNLI datasets. Each model is optimized using different estimation methods and prompt types, where GPT/Flan-T5-XXL attains the best performances in NLI capability and human alignment when using LPE/MCE. Our paper's contributions are as follows:

% Although these generative LLMs cannot capture the human disagreement distribution as small fine-tuned models, we still investigate the reason LLMs disagree.

\begin{itemize}
    \itemsep0em 
    \item To the best of our knowledge, we are the first to test generative LLMs jointly on the performance and human disagreement on NLI.
    \item We suggest two probability distribution estimation techniques for LLMs to represent disagreement and perform empirical evaluations to with respect to the human disagreement distribution.
    \item We study the model sensitivity to estimation methods and prompt types to demonstrate how these contribute to the ability of models to represent human-level disagreement for NLI.
\end{itemize}

\section{Related Works}

\subsection{Disagreement in NLI}
Considering only a single label in NLI datasets is bound to fail in capturing the diverse range of user opinions and could lead to misrepresentations of language models. To measure inherent human disagreements in NLI, \citealp{nie2020can} and \citealp{pavlick2019inherent} collected large number of human annotations (\textit{e.g.}, 100 and 50 annotations for ChaosNLI and PK2019) per instance for common NLI datasets such as SNLI \citep{bowman2015large} and MultiNLI \citep{williams-etal-2018-broad}. When taking the majority vote from these additional annotations, 22$\%$ of the instances exhibited a change in label compared to the original dataset \cite{nie2020can}.

To characterize and reproduce the extent of human disagreement in NLI tasks, previous works directly fine-tuned language models \cite{nie2020can} and implemented distribution estimation methods \cite{zhou2022distributed} using the labeled data. Other studies have constructed losses to better calibrate the ambiguity \cite{meissner2021embracing} and proposed an ensemble of models to detect disagreeing samples \cite{zhang2021identifying}.

For measuring the distance between two distributions, Kullback–Leibler (KL) Divergence \cite{kullback1951information} or its symmetric version, Jensen-Shannon Distance (JSD) \cite{endres2003new} are widely used. \citealp{baan2022stop} argued that Expected Calibration Error (ECE), the difference between the average accuracy and confidence \citep{naeini2015obtaining, guo2017calibration}, cannot capture inherent human disagreement. Therefore, for models to better calibrate to human disagreement, accuracy-agnostic metrics such as DCE have been introduced \cite{baan2022stop}.

\subsection{Alignment of Instruction-tuned LLMs }
LLMs have demonstrated the ability to follow examples provided in-context \citep{brown2020language} and have further been developed to follow natural language instructions \cite{mishra2022cross, ouyang2022training, chung2022scaling}. Instruction-following LLMs are fine-tuned with various tasks and are expected to generalize well to tasks the model was not trained on \cite{zhao2023survey}. For example, GPT-3 is fine-tuned using reinforcement learning with human feedback to produce responses that align with human values \cite{ouyang2022training}. Despite such efforts, \citealp{santurkar2023whose} % experimented with LLMs to determine whether they could reflect public opinion polls but  
identified that LLMs capture only a single perspective, exhibiting left-leaning tendencies and excluded demographic groups. Here, we study whether LLMs appropriately reflect a diversity of viewpoints in the NLI task setting.

% However, it has not yet been studied 148 whether reconstructed LLM distribution can reflect human groups in one of the basic but fundamental NLI tasks.

\section{Methods}

We estimate and quantify dissenting human voices using the multinomial soft-label distribution of LLMs with two proposed methods:

\subsection{Log Probability Estimation (LPE)} \label{sec3.1}

We use a single instance returning log probabilities of top-k\footnote{k is set to 5 for all the models to match the maximum \texttt{logprobs} size of OpenAI Completion API.} token candidates to estimate the categorical distribution of the labels. This method sums over all valid options\footnote{See Appendix \ref{app.3} for examples.}
($\mathrm{v_j}$) to estimate the model probability for class j, a method often adopted in a multiple-choice style evaluation of generative language models \cite{hendrycks2020measuring,santurkar2023whose}. Although the LPE method requires a single generation for each instance, it cannot be applied to all types of models\footnote{GPT-3.5-Turbo does not support \texttt{logprobs}.}. Additionally, the method is limited in cases where more than one token is generated as it requires exhaustive mapping of the determining token probability. Furthermore, as models only return probabilities for top-k tokens, there is an unknown non-constant probability mass. We estimate this as follows, where $C$ is the total number of classes of the task:

\begin{equation}
    \mathbf{ p(\hat{y_j}|x) \approx \frac{\sum_{i=1}^k {\exp{{{lp}_{i}}}} \cdot \mathbb{1}_{i \in v_j} }{\sum_{j=1}^C \sum_{i=1}^k {\exp{{lp_{i}}}} \cdot \mathbb{1}_{i \in v_j}} }
\end{equation}

\subsection{Monte Carlo Estimation (MCE)} \label{sec3.2}

Decoding strategies such as beam search or greedy search do not exploit the full distribution of the possible generation options. Furthermore, API-based language model services limit the number of returned token-level probabilities. Alternatively, to reconstruct the distribution of outputs from generative LLMs, we introduce an intuitive way that samples a large number\footnote{Sample size of 100 is heuristically chosen to match the size of human annotation for ChaosNLI.} of generated outputs considering the valid options\footnote{See Appendix \ref{app.3} for examples.} ($\mathrm{v_j}$) for class j. This method is based on a Monte Carlo method \cite{metropolis1949monte} to estimate the probability distribution. Even though the MCE method can be computationally expensive, it can be applied to any model and prompt type to capture the multinomial distribution of a classification setting. MCE is defined as follows:

\begin{equation}
    \mathbf{p(\hat{y_j}|x) \approx \frac{1}{n}\sum_{i=1}^n \mathbb{1}_{i \in v_j}}
\end{equation}

\begin{table*}[ht!]
\centering
\resizebox{\textwidth}{!}{%
\begin{tabular}{lccc|ccc|ccc}
\Xhline{1.0pt}
  \multicolumn{1}{c}{\multirow{2}{*}{\textbf{Model}}} &
  \multicolumn{3}{c|}{\textbf{LPE (NS)}} &
  \multicolumn{3}{c|}{\textbf{MCE (NS)}} &
  \multicolumn{3}{c}{\textbf{MCE (OS)}} \\ \cline{2-10} 
 &
  \textbf{Acc↑} &
  \textbf{JSD↓} &
  \textbf{DCE↓} &
  \textbf{Acc↑} &
  \textbf{JSD↓} &
  \textbf{DCE↓} &
  \textbf{Acc↑} &
  \textbf{JSD↓} &
  \textbf{DCE↓} \\ \Xhline{1.0pt}

\cellcolor[HTML]{FFB1B0}\textbf{Flan-T5-L} (780M)&59.3	&0.293	&0.326  &\textbf{62.3}	&\textbf{0.289}	& \textbf{0.321}    &59.7 &0.290 &0.322
 \\ 
\cellcolor[HTML]{FFB1B0}\textbf{Flan-T5-XL} (3B)&65.7	&0.253	&0.282
&\textbf{72.0}	&\textbf{0.236}	&\textbf{0.254} &70.3	&0.238	&0.256\\
\cellcolor[HTML]{FFB1B0}\textbf{Flan-T5-XXL} (11B)&68.7	&0.258	&0.277  &71.0	&0.263	&0.277  &\textbf{74.3}	&\textbf{0.232}	&\textbf{0.244}
 \\
\cellcolor[HTML]{FFDFBE}\textbf{Flan-UL2} (20B)& 67.7 & 0.260 & 0.281 &72.3	&0.247	&0.259	&\textbf{76.0}	& \textbf{0.241}	& \textbf{0.246}
 \\

\cellcolor[HTML]{FFFFBF}\textbf{OPT-IML-M-S} (1.3B)&57.0	&\textbf{0.294}	&0.337 &54.7	 &0.312	&0.354   &\textbf{59.3}	&0.298	&\textbf{0.337}
\\
\cellcolor[HTML]{FFFFBF}\textbf{OPT-IML-M-L} (30B)& 72.0	&0.273	&0.286	&62.0	&0.280	&0.303	&\textbf{72.7}	& \textbf{0.233}	& \textbf{0.252}
\\ \hline

\cellcolor[HTML]{B4F0A7}\textbf{GPT-3-D3} (175B)& 66.7	& \textbf{0.330}	& \textbf{0.345}	&\textbf{67.0}	&0.334	&0.349	&58.0	&0.344	&0.376
 \\ 
\cellcolor[HTML]{B4F0A7}\textbf{GPT-3-D2} (175B)& \textbf{64.0}	&0.282	&0.317	&62.7	& \textbf{0.279}	& \textbf{0.313}	&49.3	&0.315	&0.364 \\ \hline

\cellcolor[HTML]{A9D1F7}\textbf{Stable Vicuna} (13B)& \textbf{45.7} & \textbf{0.328} & \textbf{0.379} & 43.7 & 0.504 & 0.568 & 41.7 & 0.502 & 0.567 \\
\Xhline{1.0pt}

\end{tabular}}
\caption{Human Alignment Performances of LLMs on Subsets of ChaosNLI Datasets with Different Estimation Methods - LPE/MCE (Prompt Types - Shuffled NS/OS). We present the average results of ChaosNLI-$\alpha$, S, and M; for each dataset, we randomly sample 100 instances. The model categorizations are the same as Table \ref{tab:main}. Bold texts indicate the best value for each model and metric.}
\label{tab:tab100}
\end{table*}

\section{Experimental Design}
\subsection{Data}
\label{sec:4.1}
First, we test the inference ability of LLMs in challenging datasets, ANLI (Adversarial NLI) \cite{nie2020adversarial} and QNLI \cite{wang2018glue}. We opt for the round 3 version of ANLI (n = 1,200), which contains more contexts from diverse domains such as Wikipedia, Common Crawl, StoryCloze \cite{mostafazadeh-etal-2016-corpus}, and CBT \cite{hill2016goldilocks}. QNLI \cite{wang2018glue} (n = 5,463) is converted from the Stanford Question Answering Dataset (SQuAD) \cite{rajpurkar2016squad} to an NLI dataset, and the task is to decide whether the sentence contains the answer to the question.

Second, we jointly evaluate LLMs on ChaosNLI \cite{nie2020can}, and PK2019 \cite{pavlick2019inherent} to examine both the accuracy and how the model distribution aligns with the human disagreement distribution. These datasets consist of two task settings: ChaosNLI-$\alpha$ (n = 1,532), where models must select one of the two hypotheses, and ChaosNLI-S (n = 1,514), M (n = 1,599), and a subset\footnote{JOCI $\&$ DNC datasets of PK2019 are discarded as the annotation setting greatly varies from ChaosNLI.} of PK2019 (n = 299) where models must assign the relationship (\textit{e.g.}, entailment, contradiction, or neutral) for a pair of premise and hypothesis. We also pick out a challenging subset of the ChaosNLI datasets, which we denote as HighChaosNLI, consisting of the top 100 samples having the greatest human disagreement level.

Lastly, to trace possible causes of the disagreement occurring in LLMs, we use the round 1 version of the DisagreementNLI dataset (n = 318), where the samples from ChaosNLI are annotated with one of the 10 categories (\textit{e.g.}, probabilistic) of potential sources of disagreement \cite{jiang2022investigating}. While the primary focus is slanted towards identifying why humans disagree, we utilize and link the disagreement taxonomy to uncover whether the disagreement in LLMs aligns with those of humans.

\subsection{Models}
We categorize numerous LLMs with varying levels of supervision on the NLI task\footnote{See Appendix \ref{app.2} for more details.}: Full Exposure (FE), Partial Exposure (PE), Minimum/Unknown Exposure (MUE), and No Exposure (NE). For FE models, we follow the baseline setup of \citealp{nie2020can} by fine-tuning BERT (340M) \cite{devlin2018bert} and RoBERTa (355M) \cite{liu2019roberta}. Since instruction-following LLMs do not have full supervision of NLI, we assign these LLMs to one of the PE, MUE, and NE models.

First, the PE models include Flan-T5 (780M, 3B, 11B) \cite{chung2022scaling},  Flan-UL2 (20B) \cite{tay2022ul2}, and OPT-IML-Max (1.3B and 30B)\footnote{These models will be referred as OPT-IML-M-S and OPT-IML-M-L for convenience.} \cite{iyer2022opt}. We label  GPT-3-D2 (\verb|text-davinci-002|) and GPT-3-D3 (\verb|text-davinci-003|) (175B) \cite{ouyang2022training} as MUE models because, although the models are variants from \citealp{ouyang2022training}, it is unknown to which extent the model is exposed to NLI tasks. Finally, the sole NE model that we test is Stable Vicuna (13B) \cite{vicuna2023}\footnote{Results of poor-performing models such as Stanford Alpaca (7B) and Dolly-v2 (12B) are not reported.} because it has no exposure to NLI tasks. All the hyperparameters we use to generate the outputs of these models are listed in Appendix~\ref{app.1}.

\begin{figure}[t!] 
\centerline{\includegraphics[width=\columnwidth]{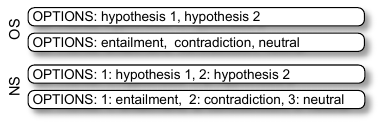}}
\caption{Two Prompt Types - OS and NS for Two Types of Tasks. The former does not have a number in front of each option choice.}
\label{fig:fig2}
\end{figure} 

\begin{table*}[ht!] 
\centering
\resizebox{\textwidth}{!}{
{%
\begin{tabular}{lcc|cccc}

\Xhline{1.0pt}

\multicolumn{1}{c}{\textbf{Model}} &
\multicolumn{1}{c}{\textbf{ANLI-R3}} &
\multicolumn{1}{c|}{\textbf{QNLI}} &
\multicolumn{1}{c}{\textbf{ChaosNLI-$\alpha$}} &
\multicolumn{1}{c}{\textbf{ChaosNLI-S}} &
\multicolumn{1}{c}{\textbf{ChaosNLI-M}} &
\multicolumn{1}{c}{\textbf{PK2019}} \\ \Xhline{1.0pt}
  
\textbf{Chance} & 33.3 & 50.0 & 50.0 & 33.3  & 33.3 & 33.3 \\ \hline

\multicolumn{7}{l}{\textit{Full Exposure (FE)}} \\ \hline 
\textbf{BERT-L$^\ast$} (340M)& 43.5 & 92.7 & 68.2 (+0.2) & 73.8 (+1.2) & 56.9 (-4.3) & - \\ 
\textbf{RoBERTa-L$^\ast$} (355M)& 44.4 & \textbf{98.9} & 83.7 (+1.6) & \underline{78.7} (+3.8) & 63.5 (-3.9) & - \\  \hline

\multicolumn{7}{l}{\textit{Partial Exposure (PE)}} \\ \hline 
\cellcolor[HTML]{FFB1B0}\textbf{Flan-T5-L} (780M)& 46.3 & 90.2 & 73.1 (+1.9) & 54.8 (-4.6) & 59.7 (+7.8) & 76.6 (+6.5)\\ 
\cellcolor[HTML]{FFB1B0}\textbf{Flan-T5-XL} (3B)& 54.3 & 93.1 & 83.3(+1.2) & 71.2 (+1.1) & 60.2 (+1.0) & \underline{76.9} (-7.4)\\ 
\cellcolor[HTML]{FFB1B0}\textbf{Flan-T5-XXL} (11B)& \textbf{58.2} & 93.7 & \underline{84.9} (+1.6) & 67.9 (+0.8) & \textbf{72.6} (+8.5) &	\textbf{82.1} (-0.6)    \\ 
\cellcolor[HTML]{FFDFBE}\textbf{Flan-UL2} (20B)& \underline{56.6} & \underline{94.9} & \textbf{86.5} (+1.8) & \textbf{79.9} (+6.4) & \underline{71.7} (+4.8) & 74.6 (-14.3)\\ 
\cellcolor[HTML]{FFFFBF}\textbf{OPT-IML-M-S} (1.3B)& 34.6 & 80.6 &53.6 (-0.7) & 66.1(+7.2) & 50.3 (+1.2) & 57.5 (-3.1)  	\\ 
\cellcolor[HTML]{FFFFBF}\textbf{OPT-IML-M-L} (30B)& 38.5 & 70.4 &72.7 (-1.8) & 77.1 (+7.3) & 65.4 (-3.5) & 68.3 (-14.5)\\ \hline 

\multicolumn{7}{l}{\textit{Minimal/Unknown Exposure (MUE)}} \\ \hline
\cellcolor[HTML]{B4F0A7}\textbf{GPT-3-D3} (175B)& 47.8 & 79.0 & 76.5 (+2.3) & 62.7 (+5.6) & 63.3 (+9.1) & 69.5 (-0.2)\\
\cellcolor[HTML]{B4F0A7}\textbf{GPT-3-D2} (175B)& 44.8 & 77.1 & 72.6 (+1.5) & 56.3 (+6.0) & 49.9 (-0.7) & 45.5 (-10.6)\\ \hline

\multicolumn{7}{l}{\textit{No Exposure (NE)}} \\ \hline 
\cellcolor[HTML]{A9D1F7}\textbf{Stable Vicuna} (13B)& 33.5 & 49.5 & 55.6 (+2.1) & 34.2 (-5.6) & 45.4 (+8.5) & 61.2 (+14.5) \\ 

% & - (-) & - (-) & - (-) & - (-)

\Xhline{1.0pt}
\end{tabular}}
}
\caption{Inference Performances of LLMs on Various Datasets. We use MCE (n = 5 for ANLI-R3/QNLI and n = 500 for ChaosNLI/PK2019) with shuffled OS. For GPTs and Stable Vicuna, we use LPE with shuffled NS. The values inside the parentheses indicate the accuracy change from the old to new labels. We report the accuracy results of the FE models ($\ast$) from \citealp{nie2020adversarial} for ANLI-R3, \citealp{devlin2018bert} and \citealp{liu2019roberta} for BERT-L and RoBERTa in QNLI, and \citealp{nie2020can} for ChaosNLI datasets. All the outputs are averaged over three runs, and bold and underlined texts indicate the first and the second best value for each column.}
\label{tab:main}
\end{table*}

\subsection{Prompt Types} \label{sec4.3}
We adopt mostly the same prompt template\footnote{Refer to Appendix \ref{app.5} for specific prompt examples.} across different types of models within each sub-dataset. Within a dataset and a model, we test two types of prompts: (1) Option Selection (OS), in which the model has to predict the name of the class label for the entailment relation, and (2) Number Selection\footnote{A multiple-choice format similar to the prompt suggested in the MMLU Benchmark \cite{hendrycks2020measuring}} (NS), in which the model has to select the number assigned to the relationship class (Figure \ref{fig:fig2}). Additionally, as LLMs are known to be sensitive to minor input modifications \citep{liang2022holistic, sun2023evaluating}, we test the effect of prompt variations over a single prompt.

NS requires the model to predict a single token of a target number and can be used with both MCE and LPE. OS, on the other hand, is not considered in the LPE formulation to encourage a scalable, comprehensive generation strategy to estimate human disagreement distribution since if we allow LPE-OS, the token-specific probability of a model output which may vary by instance/dataset/task has to be mapped per class. We implement random ordering of the options in the prompt, as also mentioned in \citealp{santurkar2023whose}, to mitigate the sensitivity due to the order of the options, which we call shuffled OS and NS throughout the paper.

% \hspace{1mm}

\subsection{Metrics}
We investigate the distribution differences between humans and LLMs at the sample level with JSD, which is a symmetric version of KL divergence \cite{endres2003new}. In addition, we evaluate human uncertainty with DCE \cite{baan2022stop} to examine how the tendencies of these two measures compare. 

\begin{gather*}
    \mathbf{JSD}(\mathbf{p}||\mathbf{q}) = \sqrt{\frac{\mathbf{KL}(\mathbf{p}||\mathbf{m}) + \mathbf{KL}(\mathbf{q}||\mathbf{m})}{2}} \\ 
    % \mathbf{KL}(\mathbf{p}||\mathbf{q}) = \sum_{i} p_i \mathrm{log}(\frac{p_i}{q_i}),\quad \mathbf{m} = \frac{\mathbf{p} + \mathbf{q}}{2}  \\
    \mathbf{DCE}(\mathbf{p}, \mathbf{q}) = \frac{1}{2} ||\mathbf{p} - \mathbf{q}||_{1} 
\end{gather*}

where $\mathbf{KL}(\mathbf{p}||\mathbf{q})=\sum_{i} p_i \mathrm{log}(\frac{p_i}{q_i})$, $\mathbf{m} = \frac{\mathbf{p} + \mathbf{q}}{2}$

\hspace{0.5mm}

\begin {figure*}[t!]
\centering
\begin{tikzpicture}
\begin{axis}[
    width  = 1.02*\textwidth,
    height = 6cm,    
    major x tick style = transparent,
    ybar=2*\pgflinewidth,
    bar width=3.0pt,
    ymajorgrids = true,
    ylabel style = {align=center},
    ylabel = \footnotesize{Distance from Human Distribution ↓},
    ymin = 0.1,
    ymax = 0.5,
    ytick={0, 0.1, 0.2, 0.3, 0.4, 0.5},
    symbolic x coords = {JSD ($\alpha$), DCE ($\alpha$), JSD (S), DCE (S), JSD (M), DCE (M), JSD (PK), DCE (PK)},
    xtick=data,
    xticklabel style = {rotate=0},
    legend columns=-1,
    legend style={
            at={(1.00, 1.01)},
            anchor=south east,
            column sep=0.5ex,
            font=\small
    },
    legend image code/.code={
        \draw [#1] (0cm,-0.1cm) rectangle (0.2cm,0.2cm); }]

    \addplot[ybar,fill=white, error bars/.cd, y dir=both, y explicit relative] coordinates {
        (JSD ($\alpha$), 0.3055) 
        (DCE ($\alpha$), 0.3188) 
        (JSD (S), 0.2300)
        (DCE (S), 0.2427)
        (JSD (M), 0.3152)
        (DCE (M), 0.3640)
        (JSD (PK), 0)
        (DCE (PK), 0)};
    \addplot[ybar, fill=white, postaction={pattern=north east lines}, error bars/.cd, y dir=both, y explicit relative] coordinates {
        (JSD ($\alpha$), 0.2128)
        (DCE ($\alpha$), 0.1906)
        (JSD (S), 0.2210)
        (DCE (S), 0.2254)
        (JSD (M), 0.3112)
        (DCE (M), 0.3507)
        (JSD (PK), 0)
        (DCE (PK), 0)};
    \addplot[ybar,fill=rred, error bars/.cd, y dir=both, y explicit relative,] coordinates {
        (JSD ($\alpha$), 0.2271) 
        (DCE ($\alpha$), 0.2542)
        (JSD (S), 0.3415) 
        (DCE (S), 0.3816)
        (JSD (M), 0.2992) 
        (DCE (M), 0.326) 
        (JSD (PK), 0.2011) 
        (DCE (PK), 0.1988) };
    \addplot[ybar,fill=rred, postaction={pattern=north east lines}, error bars/.cd, y dir=both, y explicit relative,] coordinates {
        (JSD ($\alpha$),   0.167)
        (DCE ($\alpha$),  0.1726) 
        (JSD (S),  0.2543) 
        (DCE (S),   0.2646) 
        (JSD (M),  0.2907) 
        (DCE (M),  	0.3264)
        (JSD (PK),  0.2173) 
        (DCE (PK),  0.2177) };
    \addplot[ybar,fill=rred, postaction={pattern=grid}, error bars/.cd, y dir=both, y explicit relative,] coordinates {
        (JSD ($\alpha$),  0.1585)
        (DCE ($\alpha$), 	0.1535)
        (JSD (S), 0.2704) 
        (DCE (S),  	0.2891) 
        (JSD (M),0.2702	)
        (DCE (M), 0.2917) 
        (JSD (PK),  0.226) 
        (DCE (PK), 	0.2141)};
    \addplot[ybar,fill=oorange, error bars/.cd, y dir=both, y explicit relative,] coordinates {
        (JSD ($\alpha$),   0.1556)
        (DCE ($\alpha$), 0.1452)
        (JSD (S), 0.2346) 
        (DCE (S),  0.2320)
        (JSD (M), 0.2873) 
        (DCE (M),  0.3058)
        (JSD (PK),  0.2794) 
        (DCE (PK), 0.2825)};
    \addplot[ybar,fill=yyellow, error bars/.cd, y dir=both, y explicit relative,] coordinates {
        (JSD ($\alpha$),   0.3134) 
        (DCE ($\alpha$), 0.3714)
        (JSD (S),  0.2691) 
        (DCE (S),   0.2923)
        (JSD (M),  0.3281)
        (DCE (M),  0.3713)
        (JSD (PK),  0.349) 
        (DCE (PK), 	0.393) };
    \addplot[ybar,fill=yyellow, postaction={pattern=north east lines}, error bars/.cd, y dir=both, y explicit relative,] coordinates {
        (JSD ($\alpha$),   0.2353)
        (DCE ($\alpha$),  0.2632)
        (JSD (S),  0.2287) 
        (DCE (S),  0.2486)
        (JSD (M),  0.2131) 
        (DCE (M),  0.2225) 
        (JSD (PK),  0.2704) 
        (DCE (PK),  0.2982) };
    \addplot[ybar,fill=ggreen, error bars/.cd, y dir=both, y explicit relative,] coordinates {
        (JSD ($\alpha$),  0.2543)
        (DCE ($\alpha$), 0.2488) 
        (JSD (S),  0.3477) 
        (DCE (S),  0.3735) 
        (JSD (M), 0.3732)
        (DCE (M), 0.4018) 
        (JSD (PK),   0.3209)
        (DCE (PK), 0.3329)};
    \addplot[ybar,fill=ggreen, postaction={pattern=north east lines}, error bars/.cd, y dir=both, y explicit relative,] coordinates {
        (JSD ($\alpha$),  0.2287) 
        (DCE ($\alpha$),  0.2486) 
        (JSD (S),  0.3137) 
        (DCE (S),  0.3537) 
        (JSD (M),  0.3122) 
        (DCE (M),  0.3600) 
        (JSD (PK),  0.3820) 
        (DCE (PK),  0.4391) };
    \addplot[ybar,fill=bblue, error bars/.cd, y dir=both, y explicit relative,] coordinates {
        (JSD ($\alpha$),   0.3098)
        (DCE ($\alpha$), 0.3680) 
        (JSD (S),  0.3899)
        (DCE (S),  0.4512) 
        (JSD (M),  0.3029) 
        (DCE (M),  0.3415)
        (JSD (PK),  0.2995) 
        (DCE (PK), 0.3434)};
    \addplot[ybar,fill=gray, error bars/.cd, y dir=both, y explicit relative] coordinates {
        (JSD ($\alpha$), 0.3205)
        (DCE ($\alpha$), 0.3812) 
        (JSD (S), 0.3830)
        (DCE (S), 0.4400)
        (JSD (M), 0.3023) 
        (DCE (M), 0.3443)
        (JSD (PK), 0.3491)
        (DCE (PK), 0.4268)};
    \legend {B$^\ast$, RB$^\ast$,
             FL, FXL,
             FXXL, FUL2,
             OS, OL,
             G3, G2, SV,
             Chance};
\end{axis}
\end{tikzpicture}
\caption{Human Alignment Performances of LLMs on ChaosNLI and PK2019 Datasets. The model categorizations and estimation methods are the same as Table~\ref{tab:main}. All the outputs are averaged over three runs. We additionally visualize pairwise model similarity using JSD in Appendix \ref{app.4}.}
\label{fig:fig3}
\end{figure*}
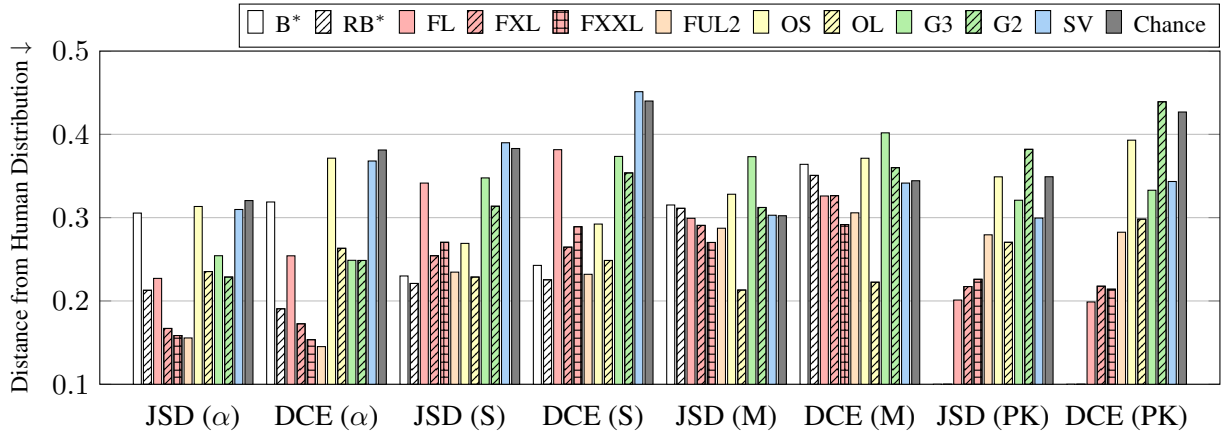

\section{Results}

\paragraph{LLMs are sensitive to different estimation methods and prompt types.} 
To select the optimal estimation methods and prompt types for each model, we examine three cases\footnote{We exclude LPE (OS) due to the reason outlined in Section \ref{sec4.3}.} - (1) LPE (NS), (2) MCE (NS), and (3) MCE (OS) for 100 randomly selected examples in ChaosNLI subsets (Table \ref{tab:tab100}). All the PE models perform the best using MCE (OS) or MCE (NS). Meanwhile, GPT-3-D3 performs better using LPE (NS) than either MCE method, hinting that larger models ($>$100B) may not need costly methods to estimate the model distribution. Similarly, for GPT-3-D2 and Stable Vicuna, a drastic negative effect of using MCE methods is exhibited, especially when using OS. Hence, we choose MCE (OS) for the PE models and LPE (NS) for the MUE and NE models.

\paragraph{The NLI capability of LLMs does not only increase due to model size.}

In Table~\ref{tab:main}, even though GPT-3-D3 has the largest parameters (175 billion) and surpasses GPT-3-D2 and Stable Vicuna, its accuracy is significantly outperformed by the PE models across ANLI-R3, QNLI, ChaosNLI, and PK2019 datasets. For ChaosNLI-S especially, GPT-3-D3 shows comparably lower performances than any FE and PE models. The leading PE models are Flan-UL2 and Flan-T5-XXL across most of the tested datasets (Table~\ref{tab:main}). The best PE model achieves 9 to 15\% higher accuracy in ANLI-R3/ChaosNLI-M than the best FE model (\textit{i.e.}, RoBERTa-L). However, Flan-T5-UL2 is marginally higher than RoBERTa-L by 1 to 3 $\%$  in ChaosNLI-$\alpha$/S, and Flan-T5-XXL even achieves 9.1\% higher than RoBERTa-L for ChaosNLI-M. Within the Flan-T5 family, scaling the model leads to enhanced inference performances. However, the largest model across all the tested models - GPT-3-D3 does not always attain the best accuracy, suggesting that model size alone is not a critical factor for performance on NLI.

\paragraph{Does multiple annotation help?}
For most of the models making inferences on the re-annotated datasets of ChaosNLI, improvements in NLI accuracy are observed, with the exception of OPT-IML-M-S. (Refer to the values inside parentheses in Table~\ref{tab:main}). This supports the necessity of having increased multiple annotations for tasks that humans are expected to disagree with. Also, it is noticeable how all these models, even if they were exposed to a sample of the train set with the original label, show better performances in the newly annotated ChaosNLI. However, we detect an accuracy decrease between the old and new labels in the PK2019 dataset for most of the models except for Flan-T5-L and Stable Vicuna. We hypothesize this is due to the way in which the final label was selected in \citealp{pavlick2019inherent}: annotators were asked to select an interval score which was later manually discretized.

\begin{figure*}[t!] 
\centerline{\includegraphics[width=\textwidth]{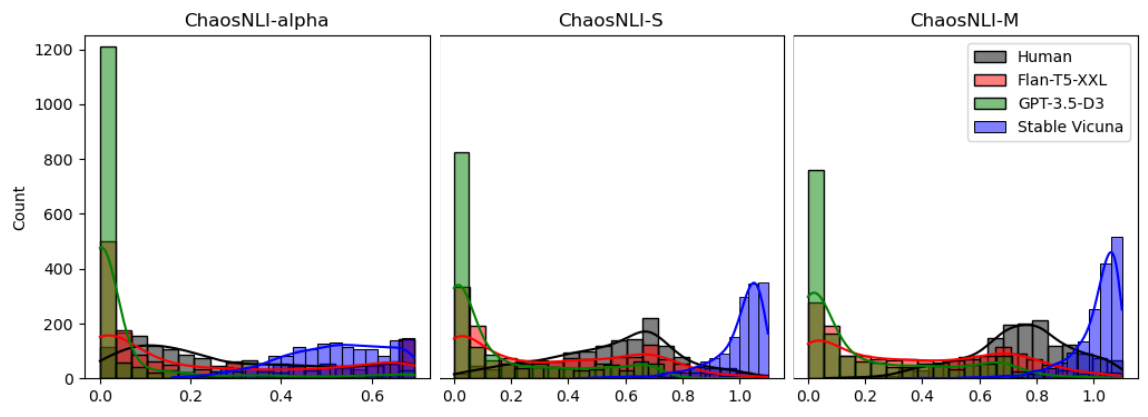}}
\caption{Histogram of Human and LLM Entropy Levels for ChaosNLI Datasets. The distributions of Flan-T5-XXL and GPT-3-D3/Stable Vicuna are estimated using MCE (OS) and LPE (NS), respectively, same as Table~\ref{tab:main}.}
\label{fig:fig4}
\end{figure*}

\paragraph{Alignment with human disagreement is not always better for larger models.}
To examine how closely the estimated distribution of LLMs aligns with the human disagreement distribution, we compare sample-level measures of JSD  and DCE between humans and LLMs (Figure~\ref{fig:fig3}). Similar to the accuracy results (Table~\ref{tab:main}), GPT-3-D3 fails to align with the human label distribution compared to some well-performing PE models, such as Flan-T5-XXL and Flan-T5-UL2. Also, each model displays a similar tendency between JSD and DCE, suggesting that either one of the metrics might be enough to measure human alignment.

As can be observed in Figure~\ref{fig:fig3}, none of the LLMs show less JSD/DCE values than RoBERTa-L in ChaosNLI-$\alpha$/S. Within LLMs, there is no one leading model that performs well across all datasets. For example, while Flan-UL2 scores the lowest JSD/DCE value in the ChaosNLI-$\alpha$ dataset, OPT-IML-M-L shows the lowest distance from human distribution in the ChaosNLI-M dataset. It is important to note that GPT-3-D3 shows worse JSD/DCE than RoBERTa-L for all ChaosNLI datasets, and it even performs worse than Stable Vicuna in ChaosNLI-M. Intriguingly, the Flan-T5 family benefits from scaling model size in ChaosNLI datasets, but Flan-T5-large does not show the highest JSD/DCE in PK2019 datasets.

\paragraph{Effect of Human Entropy on LLM Disagreement}
We filter out a challenging subset, HighChaosNLI, which is the top 100 selected samples with the highest human disagreement levels based on the entropy of each instance. We observe a plunge in accuracy as well as a rise in JSD/DCE for every model (Table~\ref{tab:high}) compared to the human alignment performances for full datasets in Table \ref{tab:main}. Still, the leading model concerning inference ability (\textit{i.e.}, Flan-T5-XXL) is unchanged, obtaining the highest accuracy of 52\% in HighChaosNLI. On the other hand, it is notable how Stable Vicuna displays the lowest JSD/DCE compared to the other models (Table~\ref{tab:high}). Nevertheless, with the hint of the worst accuracy out of all the models for full ChaosNLI datasets (Figure~\ref{fig:fig3}) and high entropy levels (Figure~\ref{fig:fig4}), we conclude that it is a mere coincidence that Stable Vicuna exhibits the best performance in terms of human alignment performances in HighChaosNLI dataset (Table~\ref{tab:high}).

\begin{table}[t!] %htbp
\centering
\resizebox{\columnwidth}{!}{
{%
\begin{tabular}{lccc}
\Xhline{1.0pt}

  \multicolumn{1}{c}{} &
  \multicolumn{3}{c}{\textbf{HighChaosNLI}} \\
  
  \cline{2-4} 

  \multicolumn{1}{c}{\multirow{-2}{*}{\textbf{Model}}} &
  \textbf{Acc↑} &
  \textbf{JSD↓} &
  \textbf{DCE↓} \\ \Xhline{1.0pt}
  
  \cellcolor[HTML]{FFB1B0}\textbf{Flan-T5-L} (780M)&44.0	&0.256	&0.318 \\ 
  \cellcolor[HTML]{FFB1B0}\textbf{Flan-T5-XL} (3B)&48.0	&0.268	&0.336 \\ 
  \cellcolor[HTML]{FFB1B0}\textbf{Flan-T5-XXL} (11B)&\textbf{52.0}	&0.300	&0.362  \\ 
  \cellcolor[HTML]{FFDFBE}\textbf{Flan-UL2} (20B)& 50.3   & 0.321    & 0.378     \\
  \cellcolor[HTML]{FFFFBF}\textbf{OPT-IML-M-S} (1.3B)&\underline{51.0}	&\underline{0.254}	&\underline{0.293}    \\ 
  \cellcolor[HTML]{FFFFBF}\textbf{OPT-IML-M-L} (30B)& 50.7   & 0.266    & 0.312   \\ \hline

  \cellcolor[HTML]{B4F0A7}\textbf{GPT-3-D3} (175B)& 50.0   & 0.435    & 0.494      \\  
  
  \cellcolor[HTML]{B4F0A7}\textbf{GPT-3-D2} (175B)& 45.7   & 0.310    & 0.354       \\ \hline
  
  \cellcolor[HTML]{A9D1F7}\textbf{Stable Vicuna} (13B)& 42.7   & \textbf{0.189}    & \textbf{0.240}     \\  \Xhline{1.0pt}
\end{tabular}}
}
\caption{Inference and Human Alignment Performances of LLMs on HighChaosNLI. The model categorizations and estimation methods are the same as Table~\ref{tab:main}. All the outputs are averaged over three runs.}
\label{tab:high}
\end{table}

We further attempt to investigate the possible causes of this phenomenon by spanning out the entropy distribution. On the consistent finding that GPT-3-D3 performs worse than Flan-T5-XXL in solving NLI tasks (Table~\ref{tab:main}) and capturing human disagreement levels (Figure~\ref{fig:fig3}), even in the HighChaosNLI dataset (Table~\ref{tab:high}), as can be observed in Figure~\ref{fig:fig4}, GPT tends to be more overconfident, showing a entropy of less than 0.1 in most samples. In contrast, the human entropy is mostly evenly distributed in the range of 0.4 to 0.6 for ChaosNLI-$\alpha$ and 0.8 to 1.0 for ChaosNLI-S/M. On the other hand, Flan-T5-XXL exhibits lower confidence than GPT-3-D3 but higher confidence than humans, and Stable Vicuna is uncertain in most instances.

\paragraph{Effect of Varying Prompts}\label{res:promptvar}
To observe the effect of prompt sensitivity on varying prompt templates, we craft variations of the pre-selected prompt. For SNLI and MNLI, we sample out five prompt variants from the Flan repository\footnote{\href{https://github.com/google-research/FLAN/}{https://github.com/google-research/FLAN/}} and make sensible variants for ChaosNLI-$\alpha$ as it is not part of the Flan mixture. From Table~\ref{tab:pv}, it is shown that the prompt variation generally benefits Flan models in the ChaosNLI-S/M datasets as they were exposed to the prompt templates. However, the pre-selected single prompt is beneficial in performance for the ChaosNLI-$\alpha$ dataset for Flan models and all datasets in GPT-3-D3 and Stable Vicuna. The performance drop using prompt variation is even more severe for GPT-3-D3, suggesting the preferred usage of a carefully crafted single prompt over using unseen input templates. However, this does not mean that the single prompt should always be preferred since variations of prompts may display fairer performance trends of diverse models within the ground of robustness.

\begin{table*}[ht!]
\centering
\resizebox{\textwidth}{!}{%
\begin{tabular}{lccc|ccc|ccc}
\Xhline{1.0pt}
  \multicolumn{1}{c}{\multirow{2}{*}{\textbf{Model}}} &
\multicolumn{3}{c|}{\textbf{ChaosNLI-$\alpha$}} &
  \multicolumn{3}{c|}{\textbf{ChaosNLI-S}} &
  \multicolumn{3}{c}{\textbf{ChaosNLI-M}} 
  \\ \cline{2-10} 
 &
  \textbf{Acc↑} &
  \textbf{JSD↓} &
  \textbf{DCE↓} &
  \textbf{Acc↑} &
  \textbf{JSD↓} &
  \textbf{DCE↓} &
  \textbf{Acc↑} &
  \textbf{JSD↓} &
  \textbf{DCE↓} \\ \Xhline{1.0pt}
\cellcolor[HTML]{FFB1B0}\textbf{Flan-T5-L} (780M)&\textbf{72.9}	&\textbf{0.228}	&\textbf{0.255}	&54.6	&0.341	&0.382	&59.7	&\textbf{0.299}	&\textbf{0.325}
 \\
 \cellcolor[HTML]{FFB1B0}{w/ Prompt Variation} &64.9	&0.279	&0.317	&\textbf{63.3}	&\textbf{0.303}	&\textbf{0.330}	&\textbf{64.7}	&0.303	&0.331
 \\ \hline
 
 \cellcolor[HTML]{FFB1B0}\textbf{Flan-T5-XL} (3B)&\textbf{83.2}	&\textbf{0.166}	&\textbf{0.171}	&71.8	&0.255	&0.264	&64.5	&0.272	&0.304
 \\
 \cellcolor[HTML]{FFB1B0}{w/ Prompt Variation} &81.6	&0.184	&0.193	&\textbf{73.8}	&\textbf{0.231}	&\textbf{0.243}	&\textbf{68.3}	&\textbf{0.271}	&\textbf{0.297}
 \\ \hline

\cellcolor[HTML]{FFB1B0}\textbf{Flan-T5-XXL} (11B)&\textbf{84.9}	&\textbf{0.159}	&\textbf{0.154}	&67.9	&0.270	&0.289	&\textbf{72.6}	&0.271	&0.293
 \\
 \cellcolor[HTML]{FFB1B0}{w/ Prompt Variation} &83.8	&0.162	&0.164	&\textbf{68.7}	&\textbf{0.259}	&\textbf{0.279}	&71.6	&\textbf{0.260}	&\textbf{0.285} 
 \\ \hline
 
\cellcolor[HTML]{B4F0A7}\textbf{GPT-3-D3} (175B)&\textbf{76.5}	&\textbf{0.254}	&\textbf{0.249}	&\textbf{62.7}	&\textbf{0.348}	&\textbf{0.374}	&\textbf{63.3}	&\textbf{0.373}	&\textbf{0.402}
 \\ 
\cellcolor[HTML]{B4F0A7}{w/ Prompt Variation} 	&72.3	&0.285	&0.29	&50.1	&0.402	&0.453	&51.5	&0.403	&0.452 
 \\  \hline
 
\cellcolor[HTML]{A9D1F7}\textbf{Stable Vicuna} (13B)&\textbf{55.6}	&\textbf{0.310}	&\textbf{0.368}	&\textbf{34.2}	&\textbf{0.390}	&\textbf{0.451}	&\textbf{45.4}	&\textbf{0.303}	&\textbf{0.342} \\
\cellcolor[HTML]{A9D1F7}{w/ Prompt Variation} 	&51.4	&0.324	&0.386	&29.9	&0.431	&0.503	&42.4	&0.337	&0.382 \\

\Xhline{1.0pt}
\end{tabular}}
\caption{Inference and Human Alignment Performances on the ChaosNLI Datasets with and without Prompt Variations. The estimation methods for each model are the same as Figure~\ref{fig:fig4}. All the outputs are averaged over three runs, and bold texts indicate the best value for each model and column.}
\label{tab:pv}
\end{table*}

\begin{figure*}[t!] 
\centerline{\includegraphics[width=\textwidth]{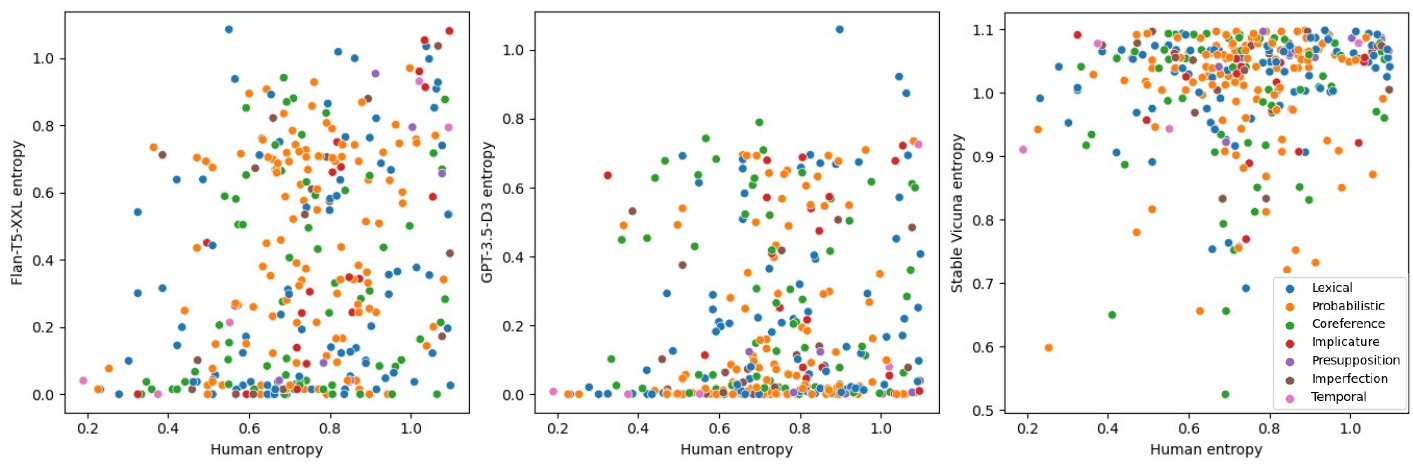}}
\caption{Relationship between Human and LLM Entropy Levels Divided with Different Human Disagreement Reasons. The estimation methods for each model are the same as Figure~\ref{fig:fig4}.}
\label{fig:fig5}
\end{figure*}

\paragraph{What causes LLMs to disagree?}
Sources of human disagreement have been well studied, but there is a lack of study of the disagreement sources for LLMs. We try to find the causes of LLM disagreements by drawing a relationship between LLM entropy level and human disagreement sources (discussed in \citealp{jiang2022investigating}) for each sample (Figure~\ref{fig:fig5}). However, no visible correlation of LLM entropy on human entropy is displayed across identified sources of human disagreement. This suggests that the cause of LLM disagreements may be due to factors other than human entropy and disagreement sources. Thus,  Under the naive assumption that LLMs will attend to similar cues to humans, we are not fully uncovering the lens of why LLMs truly disagree.

\section{Discussion}

\paragraph{LLMs do not perform well in NLI.}
Despite minimal, unknown, or absence of exposure to the NLI task, we anticipated that state-of-the-art LLMs such as GPT-3 and Stable Vicuna could reason with this relatively basic inference problem. The models are trained with billions of parameters and are known to be effective in helping real-world users solve diverse, complex tasks \cite{ouyang2022training}. However, the unforeseen poor performance of these models casts doubt as to whether they possess true general language understanding abilities. 

The problem is exacerbated for distilled models (\textit{e.g.} Stable Vicuna) that are fine-tuned using proprietary LLMs, a performance discrepancy issue similarly raised by \citealp{gudibande2023false}. Since smaller LLMs fully or partially trained with NLI tasks could perform much better than the MUE and NE models, this hints at a task-specific latent factor in NLI tasks where supervised training is beneficial and required for a wider definition of natural language understanding. In fact, as these LLMs can simply be fine-tuned to perform better for NLI tasks, a stricter evaluation criterion is needed to assess the genuine understanding capability of LLMs.

\paragraph{Characterizing Disagreement with respect to Ambiguity and Uncertainty} 

Previous studies relate multiple annotations not only to disagreement \cite{uma2021learning, gordon2021disagreement}, but also to ambiguity \cite{min2020ambigqa, tamkin2022task, liu-liu-2023-ambiguity}, and mostly to uncertainty \cite{fox2011distinguishing, xiao2021hallucination, kuhn2022semantic, zhan2023test, hu2023uncertainty}.
The definitions of ambiguity, uncertainty, and disagreement have the potential to be conflated and disambiguated. In our paper, we use the multinomial soft label estimate of a model as a representation of ``disagreement''.  When estimating this distribution with MCE, our modeling assumption treats each query to the model is analogous to asking an individual annotator to provide a label. In contrast, LPE is analogous to asking an individual to assign the scores to each option. %The multinomial \textit{disagreement} using LLM distribution reconstructed with MCE/LPE. 
Whereas most works exploit disagreement or uncertainty to improve various NLP task performances \cite{zhang2021learning, fornaciari2021beyond, yu2022learning, zhou2023navigating} our study focuses on evaluating the models. We find that using both methods for estimating the multinomial label distribution by querying the language model are not calibrated well with the human annotations.

\paragraph{Other domain tasks are transferable to NLI.}

Our work can be expanded to test LLMs on other NLP applications \cite{plank2022problem} such as Question Answering \cite{de2019commitmentbank}, Fact Verification \cite{thorne2018fever}, and Toxic Language Detection \cite{schmidt2017survey, sandri2023don}. 
Further, our method can be applied for tasks that contain disagreements since they are easily transferable to NLI tasks \cite{dagan2006pascal} like the QNLI dataset from Table~\ref{tab:main}, for example, instead of directly asking controversial questions (\textit{e.g.}, abortion) to the model \cite{santurkar2023whose}, the question format can be modified into a declarative statement in the premise and place a possible answer in the hypothesis with a binary True/False label \cite{dagan2006pascal}. Thus, if these complicated tasks can be formulated in a way where the LLM can estimate a multinomial distribution over a set of classes, our methods are applicable.

However, we should consider the target tasks when tracing “human disagreement” only when it is a significant signal that needs to be captured. For example, since it is important to include diverse opinions, we can easily apply our methods to detect disagreements in hate speech \citep{schmidt2017survey}. In contrast, spotting disagreement in the arithmetic reasoning task \cite{cobbe2021training} might be less important since it often requires a logical step-by-step reasoning procedure to obtain an accurate answer.

\paragraph{How can we better align LLMs to represent dissenting voices?}
We point out the current limitation of utilizing LLMs to represent a larger human population, especially when disagreements are present. The causes of this phenomenon are indiscernible due to the entanglement of miscalibration of out-of-distribution (OOD) inference, additional noise due to disagreement and ambiguity, prompt sensitivity, and more aspects that are yet to be identified. Even though simple remedies of temperature scaling \cite{ackley1985learning, wang2022capture}, incorporating logit bias, constrained decoding  \citep{ziems2023can}, or direct supervision to multiple annotations \citep{zhang2021learning, fornaciari-etal-2021-beyond} might mitigate the misalignment, these methods are unrealistic and not scalable due to the exhaustive hyperparameter tuning and additional data collection required to represent the population of interest.

However, as LLM applications are becoming more ubiquitous, it is important for them to faithfully represent a larger population, preferably including the voices of minorities. Thus, we suggest that future LLMs could be improved to reflect human disagreements in diverse means, for example, by fine-tuning with ambiguous instances \cite{liu2023we}. As LLMs are shown to be aware of their ignorance \cite{kadavath2022language} and have the ability to express their level of confidence \cite{lin2022teaching, zhou2023navigating}, we expect future works to address similar approaches in the aspect of alignment towards the human disagreement distribution. In this way, the reconstructed model distribution with MCE and LPE may better capture different interpretations from human individuals, aiding accountability.

% \hspace{0.5mm} 

\section{Conclusions}
In this paper, we compare the performance of instruction-following generative LLMs with other fully fine-tuned smaller models on the fundamental NLI task. First, by experimenting on four different NLI datasets, we show LLMs are not performing well in the NLI task, considering their touted language comprehension capabilities. Further, in agreement with the need for multiple annotations for disagreeable NLP tasks, LLMs also fail to align with human disagreements in the ChaosNLI and PK2019 datasets. Additional development is needed to capture representative human distributions, as well as to discover key factors to disagreement sources that can influence the LLM's answer distribution.  

\section*{Limitations}
This work shows the limited ability of billion-scale LLMs in inference and disagreement tasks. Although we test with the dataset annotated with numerous human subjects per sample, 100 people may not be enough to represent the human disagreement distribution well. After more releases of human label variation datasets, our study can be extended by covering a wider range of model types and creating evaluation benchmarks to measure the degree of disagreement. If we have robust LLMs in inference and disagreement, we could then try to find the latent factors that might not be human-interpretable but lead to disagreement in LLMs and compare them with those of humans. 

\section*{Ethics Statement}
As our work directly employs trained large language models without any extra process of fine-tuning, the risks and potential biases incurred by the model checkpoints (\textit{e.g.}, dataset selection, training configurations) remain the same as the original works. 

\section*{Acknowledgments}
This work was supported by Institute of Information \& communications Technology Planning \& Evaluation (IITP) grant funded by the Korea government (MSIT) 
(No.2019-0-00075, Artificial Intelligence Graduate School Program (KAIST)) and Artificial intelligence industrial convergence cluster development project funded by the Ministry of Science and ICT (MSIT, Korea) \& Gwangju Metropolitan City.

\bibliography{anthology,custom}
\bibliographystyle{acl_natbib}

\appendix

\section{Hyperparameters} \label{app.1}
Generally, we try to set similar hyperparameters to all the models with some exceptions due to model performance and/or cost issues.

\paragraph{Temperature}
To scale the confidence of the generated output in a post-hoc manner, we unify the temperature to be 1 (\textit{i.e.}, no scaling). There exist other precedents that use a smaller temperature for a more deterministic output \cite{santurkar2023whose} or compare outputs of models with varying temperatures \cite{ouyang2022training}. However, as we jointly assess LLMs on the accuracy of NLI and human disagreement alignment, we argue that having a fixed, un-scaled temperature to generate model outputs better aligns with our research goal of estimating model outputs to capture human disagreement distribution.

\paragraph{Generation Length}
Easily adjustable by all APIs, including OpenAPI and Huggingface, we have varying generation lengths per prompt design. As discussed in Section \ref{sec4.3}, NS is a cost-efficient alternative method for OS, solely needing a single output token of numbers. Thus in LPE, a method for single token probability output, we only use the OS prompt for effective token probability calculation. We set a maximum token output length of 10 for MCE and 1 for LPE.

\paragraph{Floating Point}
We load models of size greater than 10 billion parameters (except for GPT-3) with half the precision (bfloat16; BF16). We observe Flan-T5-XXL shows a negligible increase in performance when using the original precision (single-precision floating-point; FP32) (Table \ref{tab:tabS1}).

\begin{table}[t]
\centering
\begin{tabular}{c|c|c|c}
\Xhline{1.0pt}
\textbf{Model} & \textbf{Precision} & \textbf{Acc} & \textbf{JSD} \\ \Xhline{1.0pt} 
\multirow{2}{*}{Flan-T5-XXL} & FP32 & 75.4 & 0.233 \\ \cline{2-4} 
 & BF16 &75.1 &0.233 \\ \Xhline{1.0pt}  
\end{tabular}
\caption{Effect of Precision on Inference and Human Alignment Performances for Flan-T5-XXL. The distribution is estimated using MCE (OS), same as Table~\ref{tab:main}. The outputs are averaged over three ChaosNLI sub-datasets.}
\label{tab:tabS1}
\end{table}

\section{Levels of NLI Exposure}  \label{app.2}
We outline the level of exposure to the NLI task for each model since it is an influential factor that affects the accuracy and human-alignment performances of the models.

\subsection{Full Exposure (FE) Models}
The models below are fine-tuned with the training set of an NLI task as outlined in \citealp{nie2020can}.

% \hspace{1cm}

\begin{itemize}
    \itemsep0em 
    \item Models: BERT and RoBERTa
\end{itemize}

\subsection{Partial Exposure (PE) Models}
These models are partially exposed to the NLI task in the fine-tuning stage. However, the extent of exposure is different by the adopted fine-tuning strategy, thus listed in decreasing order.

\hspace{1cm}

\textbf{Flan Collection}
\begin{itemize}
    \itemsep0em 
    \item Models: Flan-T5 models and Flan-UL2
    \item The Flan Collection \cite{longpre2023flan} is a collection of datasets in the format of instructions to enable generalization to diverse unseen tasks. It employs a fine-tuning strategy of a maximum of 1836 NLP tasks with some NLI tasks taken into account (\textit{e.g.},  ANLI, RTE, MNLI, QNLI, SNLI, etc.).\footnote{\href{https://github.com/google-research/FLAN/tree/main}{https://github.com/google-research/FLAN/tree/main}} 

\end{itemize}

\textbf{Instruction Meta-Learning (IML) Bench}
\begin{itemize}
    \itemsep0em 
    \item Models: OPT-IML-M models
    \item Instruction Meta Learning (IML) Bench \cite{iyer2022opt} is a more common benchmark that uses 1500+ NLP tasks in the fine-tuning stage. Flan is a major portion of this benchmark, with other major portions in other large datasets. We expect some NLI exposure but not as strong as the models fine-tuned by the Flan dataset.
\end{itemize}

\subsection{Minimal/Unknown Exposure (MUE) Models}
The models below are unknown to the extent of exposure to a specific NLI task.

\begin{itemize}
    \itemsep0em 
    \item Models: GPT-3-D2, GPT-3-D3
\end{itemize}

The InstructGPT paper \cite{ouyang2022training} does elaborate that the models utilizes a reward model in the process of RLHF (Reinforcement Learning from Human Feedback), and it is fine-tuned by a variety of NLP datasets, including MNLI. However, the serviced models are not directly mapped to the models of the paper, leaving the exposure to NLI largely unknown\footnote{Refer to the \href{https://platform.openai.com/docs/model-index-for-researchers}{OpenAI Documentation}}.

\subsection{No Exposure (NE) Models}
The below model does not have any exposure to a specific NLI task.
\begin{itemize}
    \itemsep0em 
    \item Model: Stable Vicuna
\end{itemize}

% \
\section{Postprocessing}  \label{app.3}
Unlike conventional approaches of fine-tuning models directly on the downstream NLI dataset, one of the challenges in assessing an NLI task is the variability of generated outputs. To transform and choose valid options from the generated outputs, we conduct postprocessing through a manually crafted dictionary for each option (\textit{See} Valid option examples on the last page.).

\section{Distribution Alignment Among LLMs}  \label{app.4}
\begin{figure*}[t!]
\centerline{\includegraphics[width=16cm]{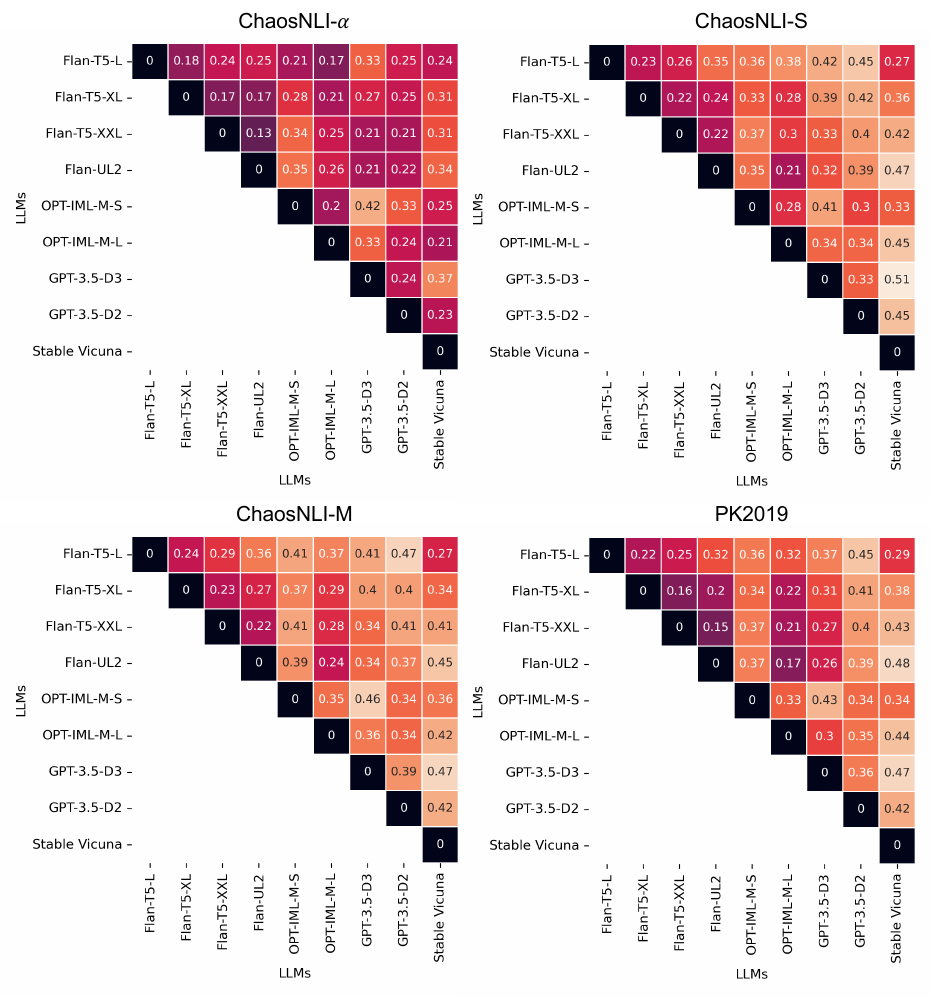}}
\caption{JSD Distribution between All Combinations of Pairs for LLMs. The darker plot indicates a similar distribution between a pair of models. The estimation methods for each model are the same as Table~\ref{tab:main}.}
\label{fig:fig6}
\end{figure*}

We illustrate the averaged sample-level JSD entropy for each model pair (Figure~\ref{fig:fig6}) to visualize the trend of alignment among LLMs. Throughout all four JSD distribution plots, the scale and range of the JSD values differ for each data. Still, the general trend is maintained, where ChaosNLI-$\alpha$ shows low JSD values overall, likely attributed to lower task difficulty witnessed by the performance gap among datasets in Table~\ref{tab:main}. The best-performing models, Flan-T5-XXL and Flan-UL2, present the lowest disagreement in entropy for all plots. 

Although the size and type of model are influencing factors, the most consistent factor is the type of instruction fine-tuning introduced for each model. Throughout all plots, the alignment is well shown for the group of models fine-tuned by the Flan dataset and the IML Bench. As we expect more research in the scope of human alignment in NLP, the evaluation of the human alignment among the models with the same fine-tuning process can also be studied and reported.

However, a strong distinction needs to be made in which an overall lower number of JSD values in this plot does not mean that a model has always had a good performance in human disagreement alignment. This figure merely delineates the alignment trends among models.

\section{Effect of Few-Shot Examples}
\begin{table*}[ht!]
\centering
\resizebox{\textwidth}{!}{%
\begin{tabular}{lccc|ccc|ccc}
\Xhline{1.0pt}
  \multicolumn{1}{c}{\multirow{2}{*}{\textbf{Model}}} &
\multicolumn{3}{c|}{\textbf{ChaosNLI-$\alpha$}} &
  \multicolumn{3}{c|}{\textbf{ChaosNLI-S}} &
  \multicolumn{3}{c}{\textbf{ChaosNLI-M}} 
  \\ \cline{2-10} 
 &
  \textbf{Acc↑} &
  \textbf{JSD↓} &
  \textbf{DCE↓} &
  \textbf{Acc↑} &
  \textbf{JSD↓} &
  \textbf{DCE↓} &
  \textbf{Acc↑} &
  \textbf{JSD↓} &
  \textbf{DCE↓} \\ \Xhline{1.0pt}
\cellcolor[HTML]{FFB1B0}\textbf{Flan-T5-XXL (0 Shot)} &\textbf{85.0}	&0.160	&0.155	&67.3	&\textbf{0.271}	&\textbf{0.291}	&72.6	&0.269	&0.290 \\
\cellcolor[HTML]{FFB1B0}\textbf{+ 1 Shot} &84.9	&0.159	&\textbf{0.154}	&\textbf{68.0}	&0.278	&0.296	&\textbf{74.8}	&\textbf{0.261}	&\textbf{0.278} \\
\cellcolor[HTML]{FFB1B0}\textbf{+ 3 Shot} 	&83.6	&0.163	&0.160	&67.0	&0.285	&0.304	&74.0	&0.271	&0.290 \\
\cellcolor[HTML]{FFB1B0}\textbf{+ 5 Shot} 	&84.9	&\textbf{0.158}	&\textbf{0.154}	&65.7	&0.288	&0.309	&73.2	&0.275	&0.295 \\ \hline

\cellcolor[HTML]{B4F0A7}\textbf{GPT-3-D3 (0 Shot)} &76.1	&0.254	&0.249	&\textbf{62.4}	&\textbf{0.348}	&\textbf{0.374}	&\textbf{63.0}	&\textbf{0.376}	&\textbf{0.405} \\ 
\cellcolor[HTML]{B4F0A7}\textbf{+ 1 Shot}  &77.7	&0.240	&0.235	&6.2	&0.400	&0.433	&61.1	&0.414	&0.445\\
\cellcolor[HTML]{B4F0A7}\textbf{+ 3 Shot}&80.7	&\textbf{0.233}	&0.216	&55.7	&0.407	&0.442	&58.2	&0.436	&0.471 \\
\cellcolor[HTML]{B4F0A7}\textbf{+ 5 Shot}  &\textbf{81.7}	&\textbf{0.233}	&\textbf{0.213}	&57.9	&0.396	&0.426	&62.2	&0.425	&0.454\\ \hline

\cellcolor[HTML]{A9D1F7}\textbf{Stable Vicuna (0 Shot)}  &55.7	&0.310	&0.368	&33.5	&0.391	&0.454	&\textbf{46.2}	&\textbf{0.304}	&\textbf{0.342} \\
\cellcolor[HTML]{A9D1F7}\textbf{+ 1 Shot} &\textbf{64.3} &\textbf{0.290}	&\textbf{0.336}	&38.6	&0.377	&0.436	&41.7	&0.311	&0.355\\
\cellcolor[HTML]{A9D1F7}\textbf{+ 3 Shot} &56.2 &0.296	&0.346	&\textbf{43.4}	&\textbf{0.351}	&\textbf{0.405}	&32.8	&0.352	&0.418 \\
\cellcolor[HTML]{A9D1F7}\textbf{+ 5 Shot} &56.1 &0.298	&0.350	&35.8	&0.379	&0.441	&28.8	&0.370	&0.441 \\
\Xhline{1.0pt}
\end{tabular}}
\caption{Inference and Human Alignment Performances on the ChaosNLI Datasets for Zero-shot and Few-shot Settings. The estimation methods for each model are the same as Table~\ref{tab:main}. Bold texts indicate the best value for each model and column.}
\label{tab:fs}
\end{table*}

We observe no consistent benefit nor harm in experimenting with few-shot settings that resemble the human annotation process more than zero-shot settings (Table~\ref{tab:fs}). In fact, zero-shot evaluation generally seems to show better performances across datasets and models compared to the few-shot evaluations. In the case of Stable Vicuna, the performance increases in the 1-shot setting for $\alpha$-NLI and the 3-shot setting for SNLI. However, we notice a plunge in 5-shot performance, especially for the MNLI dataset.

\section{Prompt Examples}  \label{app.5}
We present examples of prompts we used during the generation process in Figure~\ref{fig:fig1} (\textit{See} two prompt examples on the last page.). We incorporate a suggested general prompt template pre-specified for a specific model. For example, we implement a human and assistant-style prompt template for Stable Vicuna. Otherwise, we leave the template format the same for the rest of the models.

\clearpage
\begin{table*}[t!]
\begin{small}
\begin{tcolorbox}
[toprule=1mm,colback=black!5!white,colframe=black!75!white,title= Valid option examples for ChaosNLI-$\alpha$/S/M and two prompt types - OS and NS, fonttitle=\bfseries\large]
\begin{verbatim}
dict_alphanli_OS = {'1' : ['1','Hypothesis 1',...]
                    '2' : ['2','Hypothesis 2',...]}

dict_alphanli_NS = {'1' : '1'
                    '2' : '2'}      
                    
dict_s&mnli_OS = {'e' : ['entail','infer','yes', ...]
                  'c' : ['contradict','oppose','no', ...]
                  'n' : ['neutral','unanswerable',...]}
            
dict_s&mnli_NS = {'e' : '1'
                  'c' : '2'
                  'n' : '3'}
\end{verbatim}
\end{tcolorbox}
\end{small}
\end{table*}

\begin{table*}[htbp]
\begin{tcolorbox}[toprule=1mm,colback=black!5!white,colframe=black!75!white,title=Prompt example for ChaosNLI-$\alpha$ using OS, fonttitle=\bfseries\large]
\textbf{INPUT} \\

Read the following and determine if the hypothesis can be inferred from the premise. \\
Observation Start: My roommates put up their Christmas tree this year. \\
Observation End: This is what it's like living with a cat. \\
Hypothesis 1: The roommates soon had to take the tree down. \\
Hypothesis 2: The cat enjoyed the ornaments and garland and slept under the tree. \\
Options: Hypothesis 1, Hypothesis 2 

\tcblower 
\textbf{OUTPUT} \\ 

Answer: <Generated Output>
\end{tcolorbox}
\end{table*}

\begin{table*}[htbp]
\begin{tcolorbox}
[toprule=1mm,colback=black!5!white,colframe=black!75!white,title=Prompt example for ChaosNLI-S/M using NS, fonttitle=\bfseries\large]
\textbf{INPUT} \\

Read the following and determine if the hypothesis can be inferred from the premise. \\
Premise: This town, which flourished between 6500 and 5500 b.c. ... appear on Anatolian kilims. \\
Hypothesis: This town is over 8000 years old. \\
Options: 1: entailment, 2: contradiction, 3: neutral  

\tcblower
\textbf{OUTPUT} \\

Answer: <Generated Output>
\end{tcolorbox}
\end{table*}

\end{document}